\documentclass{article}
\usepackage[utf8]{inputenc}
\usepackage{fullpage}
\usepackage{amsmath}
\usepackage{url}
\usepackage{graphicx}
\usepackage{graphics}
\usepackage{booktabs}
\usepackage[style=alphabetic,backend=bibtex,maxnames=12,maxbibnames=10,maxcitenames=10,maxalphanames=10,giveninits=true,doi=false,url=true]{biblatex}
\newcommand*{\citet}[1]{\AtNextCite{\AtEachCitekey{\defcounter{maxnames}{2}}} \textcite{#1}}

\newcommand*{\citep}[1]{\cite{#1}}

\bibliography{freeprivacy}

\usepackage{subcaption}
\usepackage{caption}
\title{No Free Lunch in ``Privacy for Free: How does Dataset Condensation Help Privacy''}
\author{Nicholas Carlini \\ Google \and Vitaly Feldman \\ Apple \and Milad Nasr \\ Google}
\date{}
\usepackage{xcolor}
\newcommand{\eps}{\varepsilon}

\iftrue
\newcommand{\milad}[1]{\textcolor{cyan}{Milad: #1}}
\newcommand{\vitaly}[1]{\textcolor{red}{Vitaly: #1}}
\newcommand{\todo}[1]{\textcolor{green}{To do: #1}}
\else
\newcommand{\milad}[1]{}
\newcommand{\vitaly}[1]{}
\newcommand{\todo}[1]{}
\fi

\begin{document}
\maketitle
\begin{abstract}
    New methods designed to preserve data privacy require careful scrutiny. Failure to preserve privacy is hard to detect, and yet can lead to catastrophic results when a system implementing a ``privacy-preserving'' method is attacked. A recent work selected for an Outstanding Paper Award at ICML 2022 \citep{dong2022privacy} claims that dataset condensation (DC) significantly improves data privacy when training machine learning models. This claim is supported by theoretical analysis of a specific dataset condensation technique and an empirical evaluation of resistance to some existing membership inference attacks.

    In this note we examine the claims in \citep{dong2022privacy} and describe major flaws in the empirical evaluation of the method and its theoretical analysis. These flaws imply that \citep{dong2022privacy} does not provide statistically significant evidence that DC improves the privacy of training ML models over a naive baseline. Moreover, previously published results show that DP-SGD, the standard approach to privacy preserving ML, simultaneously gives better accuracy and achieves a (provably) lower membership attack success rate.
\end{abstract}

\section{Flaws in Experimental Evidence}

In the problem of dataset condensation (DC) the input is a large dataset $T = \{x_1,\ldots,x_n\}$. The goal is to output a small dataset $S = \{s_1,\ldots, s_m\}$, where $m = r_{ipc} \cdot n$  for some fraction $r_{ipc} \ll 1$ (e.g.~$r_{ipc} = 1/100$) that is ``almost'' as good as $T$ when used for training a learning algorithm. More formally, the expected generalization error when training a model on $S$ should be comparable to expected generalization error when training a model on $T$.

In \citep{dong2022privacy} the authors propose to use DC to improve data privacy of training a machine learning model. The primary focus of this work is the distribution matching (DM) technique of dataset condensation \cite{zhaobilen21}.

\subsection{The comparison to the naive baseline is incorrect}

\cite{dong2022privacy} compare the privacy of DM and several other dataset condensation schemes to the privacy of
a much simpler condensation scheme: simply choose a random subset of the training data as the condensed dataset.
When running this baseline, however, the paper incorrectly measures the attack advantage and the corresponding ROC curve.
As a result, the reported baseline attack advantage rate of $92.8\%$ is wrong:
the correct figure is just $1.6\%$.
This implies that DM's $1.06 \pm 1.20\%$ attack advantage rate does not give a statistically significant advantage over the baseline.

We begin with some notation. Let $\mathcal{A}_{\text{condense}} : T \to f$ be an algorithm that takes a full training dataset $T$, then condenses it to a dataset $S$ via a  condensation algorithm, and then trains a model $f$ on this smaller dataset $S$.
The proposed baseline considers a trivial condensation algorithm $\mathcal{A}_{\text{random}} : T \to f$ that takes a full training dataset $T$,
``condenses'' it to a dataset $S$ by randomly sampling $1\%$\footnote{The values ($r_{ipc} = 0.02$ and $r_{ipc} = 0.002$) are evaluated in \cite{dong2022privacy} but we discuss only the results $1\%$ case for brevity.} of the dataset $T$, and then trains a model $f$ on $S$.

Recall what it means to measure the accuracy of an adversary who aims to perform a membership inference attack (MIA)~\cite{shokri2017membership,salem2018ml,nasr2019comprehensive,sablayrolles2019white,jagielski2020auditing,nasr2021adversary,song2021systematic,carlini2022membership}.
Summarized briefly, we should perform the following steps:
\begin{enumerate}
    \item The attacker and defender agree on a ``universe'' of possible samples $U$.
    For this note, $U$ is the entire training data of CIFAR-10.
    \item The defender randomly samples a new dataset $T \subset U$ by taking any $x \in U$ with $50\%$ probability.\footnote{While in general the sampling probability can be arbitrary, \cite{dong2022privacy} use 50\%.}
    \item Then, the defender trains a model $f$ on $T$
    and sends $f$ to the attacker. (In \cite{dong2022privacy} the attacker is allowed access to intermediate computations used to produce $f$ but neither their no our attacks make use of anything besides $f$.)
    \item The attacker runs a membership inference attack on each example $x \in U$.
    The attacker scores 1 point if they correctly guess the membership status (i.e., the attacker guesses ``member'' and $x \in T$ or the attacker guesses ``nonmember'' and $x \not\in T$); if the guess is incorrect the attacker scores 0 points.
    \item The attack success rate is then computed the average of this score over each example in the dataset $U$. From here we can compute the \textbf{attack advantage} as $2 \times (\mbox{success rate} - 50\%)$, i.e., how much better the attack performs than random chance.
\end{enumerate}

\begin{figure}
    \centering
    \includegraphics[scale=0.4]{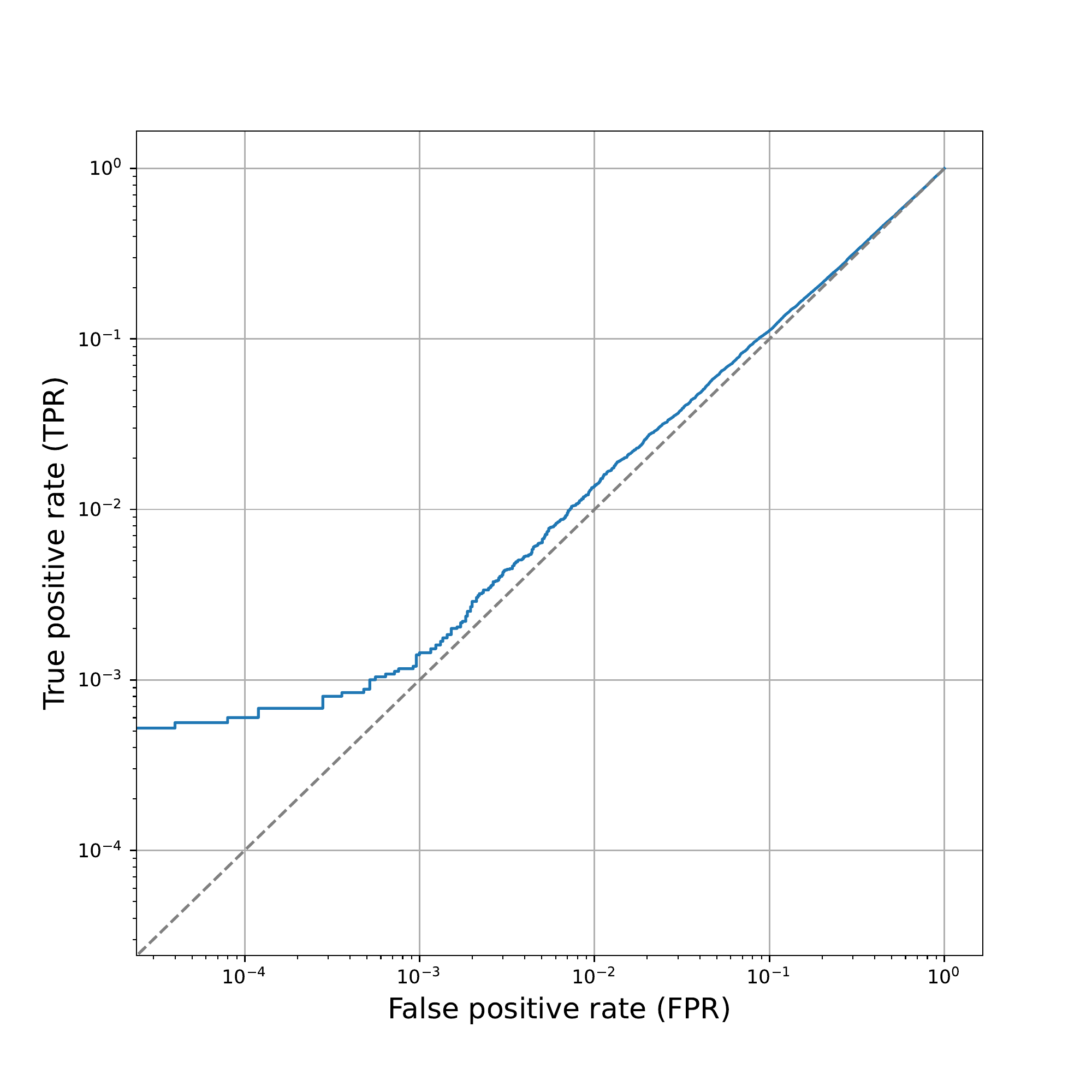}
    \caption{Membership inference accuracy of a corrected baseline compression technique on CIFAR-10, measured over the full dataset.}
    \label{fig:diff_threatmodel}
\end{figure}

This is how \cite{dong2022privacy} evaluate the privacy of their proposed scheme when evaluating DM.
But this is \textbf{not} how \cite{dong2022privacy} evaluate the privacy of the baseline.

Specifically, when \cite{dong2022privacy} evaluate the privacy of the baseline, the accuracy is measured on a set $S \cup S'$ where $S$ is the output of the baseline defense (i.e., a randomly selected $1\%$ subset of $T$), but where $S'$ is a non-member set of the same size randomly selected from $T\setminus S$.
This is not correct.
It is unfair to compare one scheme over a full universe $U$ to another scheme that evaluates over a (much smaller and more ``vulnerable''!) subset $S \cup S'$.
Viewed differently, \emph{attack accuracy} is strongly influenced by the overall base rate;
when the base rate of members is 1\%, the attack is much harder than when the
base rate is 50\%.
And \cite{dong2022privacy} here is comparing one attack with a base rate of 1\% to
another attack with a base rate of $50\%$.

When we re-evaluate this baseline correctly---by correctly following the protocol---\textbf{we obtain an attack advantage of just 1.6\%, over fifty times weaker than the claim} of $92.8\pm 5.31 \%$.
Figure~\ref{fig:diff_threatmodel} shows the ROC curve of our proper baseline, and as we can see while the adversary can still distinguish which instances are sub-sampled, the adversary cannot determine which instance were member of the training dataset but not selected to be trained on. We also remark that the highest expected MIA advantage that can be attained when only $1\%$ of the members dataset is used for training is $2\%$.

For comparison, MIA accuracy of DM on CIFAR-10 as stated \cite{dong2022privacy} is  $1.06 \pm 1.20\%$. Thus DM does not provide a statistically significant improvement over the baseline invalidating the main claim of this work.

We note that \cite[p.7]{dong2022privacy} acknowledge the difference in the evaluation setups as ``we vary a little bit the attack setting''. They also evaluate MIA of DM initialized with a random subset $S$ and attack restricted to $S \cup S'$. As noted by the authors, MIA accuracy of DM in this setting is comparable to that of the baseline further suggesting that DM does not have an advantage over the baseline.

\subsection{DM does not offer better privacy-utility tradeoff than published baselines}

\begin{table}[]
    \centering
    \caption{The DM scheme \cite{dong2022privacy} is strictly dominated by DP-SGD \cite{abadi2016deep} along every dimension. We used $\delta=10^{-5}$ in the privacy analysis.}
    \begin{tabular}{|l|c|r|r|r|r}
    \toprule
        Technique & Test Accuracy & DP $\varepsilon$ & Formal Gurantees & Attack Advantage \\
        \midrule
        DM \cite{dong2022privacy} & 59\% & N/A & No & 1.06 $\pm$ 1.2\% \\
        DP-SGD (Same arch as in~\cite{dong2022privacy}) & 61\% & 8 & Yes & 2.8 $\pm$ 0.8\% \\
        DP-SGD (Same arch as in~\cite{dong2022privacy}) & 77\% & $>$5000 & No & 3.0 $\pm$ 1.0\% \\
        \midrule
        DP-SGD~\cite{tramer2020differentially} & 64$\%$ & 2 & Yes & $<$0.8\% \\
        DP-SGD~\cite{de2022unlocking} & 66\% & 2 & Yes & $<$0.8\% \\
        \bottomrule
    \end{tabular}
    \label{tab:summary}
\end{table}

It is claimed that DM provides privacy ``for free''. This is not the case.
First, note that DM achieves significantly lower accuracy than models trained on the entire dataset. For example, DM on CIFAR-10 with $r_{ipc}=0.01$ achieves an accuracy of $59\%$ compared to state-of-the-art models that reach well over $96\%$ accuracy---so any privacy that is offered does cost at least $37\%$ drop in accuracy.

But more importantly, previously published privacy-preserving defenses \cite{de2022unlocking,tramer2020differentially} already have trained models that achieve \emph{higher accuracy} while \emph{also providing better MIA protection and provable privacy guarantees}.
\cite{tramer2020differentially} achieve $64\%$ accuracy ($5\%$ higher accuracy)
at the very conservative value of
$\varepsilon=2$ (better provable privacy than the claimed ``empirical'' bound of $\hat\eps = 2.3$),
while also maintaining an identical attack success rate (within the margin of error). A more recent work~\cite{de2022unlocking} also showed it is possible to achieve high accuracy with the same privacy budget.

\begin{figure}
    \centering
    \includegraphics[scale=0.4]{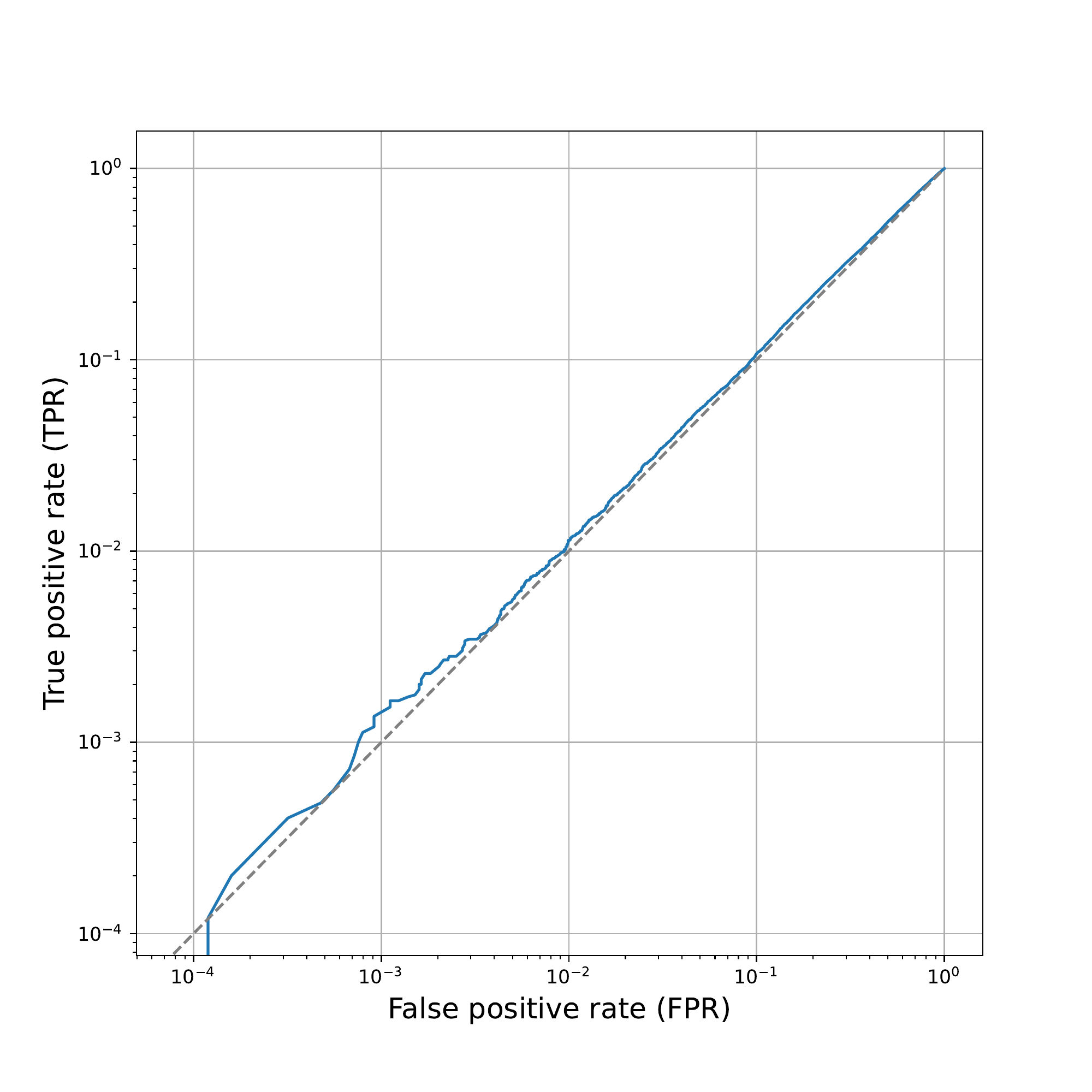}
    \caption{Training CIFAR-10 dataset by DP-SGD using the ConvNet architecture (same exact network as in \citep{dong2022privacy}) and $\varepsilon=8$.}
    \label{fig:cifar10_dpsgd_eps8}
\end{figure}

To be even more direct in our comparison,
we also train two \emph{new} models using DP-SGD with the same network architecture as \cite{dong2022privacy} (even though it is not optimized for the use of DP-SGD).
Compared to DM's $59\%$ accuracy and $1.06 \pm 1.2\%$ membership inference attack advantage,
our model achieves $77\%$ accuracy ($18\%$ higher than DM) and $3.0 \pm 0.8\%$
attack  advantage (again within the margin of error).
If we want to be even more conservative, we can train the same model used in DM to $\varepsilon=8$ differential privacy which \emph{still} outperforms the accuracy of DM:
$61\%$ accuracy (2\% better) and $2.8\%$ attack advantage (again within the margin of error).
Figure~\ref{fig:cifar10_dpsgd_eps8} shows the membership inference attack ROC curve for this settings, also the membership inference attack advantage is $2.8\%$.

Table~\ref{tab:summary} summarizes these results, and as we can see, DM is never superior along any dimension.

\subsection{The empirical $\hat\eps$ is misleading}
To relate the results to differential privacy and to compare against differentially private dataset generation techniques \cite{dong2022privacy} claim  an empirical value of $\hat\eps=2.30$ for their algorithm. This approach appears to be based on prior work that establishes empirical \emph{lower} bounds on the differential privacy parameter $\eps$~\cite{jagielski2020auditing,nasr2021adversary}.
Unlike in prior work, \cite{dong2022privacy} use the value they calculate as a privacy guarantee, that is an \emph{upper} bound on a privacy parameter. However \cite{dong2022privacy} do not formally define the value they estimate and the specific $\hat\eps$ is neither an upper bound nor a lower bound on the differential privacy parameter. 

The value $\hat\eps$ is computed by running a state-of-the-art membership inference \cite{carlini2022membership}
attack and reading off one particular true positive to false positive ratio along this curve.
Because $\ln(\frac{\text{TPR}}{\text{FPR}})$ lower bounds $\eps$,
by running an attack and choosing the maximum ratio between these two
quantities it is possible to lower bound DP parameter $\eps$.
However, the use of $\hat\eps$ computed in this way as a privacy guarantee has at least two flaws.
While it is true that $\ln(\frac{\text{TPR}}{\text{FPR}})$ lower bounds DP $\eps$,
it is invalid to run an attack once and read a single point off the ROC curve
and use the true positive and false positive rate at this single point because we have \emph{no statistical confidence} in the results.
%
%
%

Statistical techniques need to be used to ensure we make claims that are correct instead of measuring potential flukes.
In particular, when prior work has reported results by studying the TPR-FPR tradeoff, it has used a threshold of $95\%$ confidence~\cite{jagielski2020auditing,nasr2021adversary,de2022unlocking}.

The second flaw is that \emph{average-case measurements can be misleading when used as a privacy guarantee}.
A value of  $\hat\eps = 0.2$ (corresponding to TPR/FPR ratio of $\approx 1.22$) over the entire universe dataset can correspond to TPR/FPR ratio of $\infty$ over a specific identifiable subgroup comprising $10\%$ of the dataset (and ratio $1$ over the remainder of the dataset). The situation can be even more extreme when the attacker is only interested in the membership of a very specific individual in the dataset. In contrast, differential privacy limits the ability to infer presence of any individual in any dataset. For this reason, when prior work has established empirical lower bounds on the value of DP
$\eps$ they have used specific \emph{auditing} techniques~\cite{jagielski2020auditing,nasr2021adversary} that attempt to reason over worst-case datasets---not just average-case datasets.
%

To show the difference between an average case and a more adversarial case we designed a simple auditing scenario.
We construct two datasets, $D$ and $D'$, each of which is a modified version of the CIFAR-10 dataset where we removed all of the examples from class zero except one instance for dataset $D$ and two instances for dataset $D'$. As a result when we do dataset condensation on dataset $D$ all of the condensed examples of class zero are the only example for class zero, but for dataset $D'$ condense examples contain information about both examples. Therefore, when train a classifier on these condense examples and we query the second example from the dataset $D'$ it can reveal if the classifier model was trained on that example or not.
Note that these two datasets differ by exactly one example, and so any algorithm that satisfies differential privacy should produce an indistinguishable output.

We find that DM does not hide the presence of an extra point on this adversarially crafted  dataset.
To demonstrate this, we run the DM training pipeline\footnote{We used Distribution matching (DM) approach with random initialization and \texttt{color\_crop\_cutout\_flip\_scale\_rotate} for augmentations, we used the same number of condensed images for both $D$ and $D'$} to train many models on either $D$ or $D'$ and ask an adversary to distinguish between the two.
Using this simple worst-case setting we a achieve $100\%$ detection rate at distinguishing between $D$ and $D'$.
%

\section{Issues with the model and theoretical analysis of privacy}
\citep{dong2022privacy} aim to demonstrate that formal privacy guarantees can be established for the distribution matching (DM) technique of dataset condensation \cite{zhaobilen21}. For this purpose they consider a special case of the technique restricted to linear feature extractors. However, we believe that the dataset condensation algorithm resulting from this restriction cannot give any nontrivial results. In addition, the privacy analysis is based on an unrealistic assumption that itself implies differential privacy.

\subsection{Model}
The theoretical analysis in this work is for the distribution matching technique of dataset condensation \cite{zhaobilen21}.
Proposition 4.4 \citep{dong2022privacy} characterizes the technique (for linear feature extractors) the rest of the subsequent privacy analysis is applied to this characterization. Specifically, this characterization states the following. If the technique is initialized\footnote{In Proposition 4.4 the initial dataset is referred to as $S$ and the resulting condensed dataset $S^*$. This is inconsistent with the notation in the rest of the work so we a different notation.} with a set of examples $S_{\mbox{init}}=\{s'_1,\ldots,s'_m\}$ and run on the training dataset  $T=\{x_1,\ldots,x_n\}$ then it will result in a set of examples $S = \{s_1,\ldots,s_m\}$ such that for all $i \in [m]$,
$$s_i = s'_i - \frac{1}{m}\sum_{j\in [m]} s'_j + \frac{1}{n}\sum_{j\in [n]} x_j .$$
Note that this transformation amounts to centering the dataset at the mean of the original dataset (instead of the mean of $S_{\mbox{init}}$). In the first initialization technique that  \citep{dong2022privacy}  consider the dataset $S_{\mbox{init}}$ is chosen to be a random subset of $T$ then. In this case, by the standard high-dimensional concentration results, $$\frac{1}{m}\sum_{j\in [m]} s'_j \approx \frac{1}{n}\sum_{j\in [n]} x_j$$ and the transformation is essentially an identity.

The second initialization scheme this work considers is sampling from the $d$-dimensional normal distribution subject to the mean of the samples being the origin (in particular, $S_{\mbox{init}}$ is centered). DM applied to this dataset will result in a dataset which is centered at the mean of $T$ but is otherwise independent of $T$. In particular, if the dataset $T$ already happens to be centered (which is a common preprocessing step) then DM is just an identity transformation.

It is clear that such a transformation cannot possibly lead to a perceptible improvement in the generalization error of a learning algorithm over running the algorithm on $S_{\mbox{init}}$ itself. Thus the theoretical analysis in \citep{dong2022privacy} is uninformative.

\subsection{Privacy analysis}

The privacy analysis is based on the Assumption~4.8 \citep{dong2022privacy} stating that given the training set $T$, a learning algorithm outputs a model $\theta$, from a distribution proportional to $\exp(-\sum_{i\in[n]} l(\theta,x_i))$, where $l$ is the loss function.
This is the standard exponential mechanism that is well-known\footnote{\url{ https://en.wikipedia.org/wiki/Exponential_mechanism_(differential_privacy)}} to be $\epsilon$-DP where $\epsilon$ is equal to twice the range of the loss function $l(\theta,\cdot)$. Naturally, had the model been output from such a distribution we would have privacy and membership inference would not achieve high-accuracy (with reasonable assumptions on the range of the loss such as the one made in Assumption~4.9 of \citep{dong2022privacy}).

The assumption is borrowed from \citep{sablayrolles2019white}, where it is used to prove that, under this assumption black-box membership inference is as accurate as white-box and to design practical membership inference attacks. In particular, the use is ``harmless'' as it is not used to provide privacy guarantees.

Finally, we note that it should not be surprising or useful to know that an algorithm that uses little information about $T$ to modify $S$ (just the mean of $T$) is private on points in $T\setminus S$.

\section*{Acknowledgements}
We thank Gautam Kamath, Aleksandar Nikolov, Thomas Steinke, Kunal Talwar and Florian Tramer for their valuable commments and useful suggestions on this note.

\printbibliography
\end{document}